\documentclass[runningheads]{llncs}
\usepackage[T1]{fontenc}
\usepackage{booktabs}
\usepackage{multirow}
\usepackage{array}
\usepackage{amssymb}
\usepackage{amsmath}
\usepackage{marvosym}
\usepackage{hyperref}
\usepackage{makecell}
\usepackage{multirow}
\usepackage{graphicx}
\usepackage{xcolor}
\usepackage{subcaption}
\usepackage{colortbl}
\definecolor{mygray}{gray}{.9}
\begin{document}
\title{SimTxtSeg: Weakly-Supervised Medical Image Segmentation with Simple Text Cues}
%
%
\author{Yuxin Xie\inst{1,2} \and
Tao Zhou\inst{3} \and
Yi Zhou\inst{1,2}\thanks{Corresponding author: Yi Zhou} \and
Geng Chen\inst{4}}
%
\authorrunning{Y. Xie and Y. Zhou et al.}
%
\institute{School of Computer Science and Engineering, Southeast University, China \and Key Laboratory of New Generation Artificial Intelligence Technology and Its Interdisciplinary Applications, Ministry of Education, China \and Nanjing University of Science and Technology, China \and
Northwestern Polytechnical University, China \\
\email{\{silver\_iris@163.com, yizhou.szcn@gmail.com\}}
}
\maketitle              
\begin{abstract}
Weakly-supervised medical image segmentation is a challenging task that aims to reduce the annotation cost while keep the segmentation performance. In this paper, we present a novel framework, SimTxtSeg, that leverages simple text cues to generate high-quality pseudo-labels and study the cross-modal fusion in training segmentation models, simultaneously. Our contribution consists of two key components: an effective Textual-to-Visual Cue Converter that produces visual prompts from text prompts on medical images, and a text-guided segmentation model with Text-Vision Hybrid Attention that fuses text and image features. We evaluate our framework on two medical image segmentation tasks: colonic polyp segmentation and MRI brain tumor segmentation, and achieve consistent state-of-the-art performance. Source code is available at: \href{https://github.com/xyx1024/SimTxtSeg}{https://github.com/xyx1024/SimTxtSeg}.

\end{abstract}
\section{Introduction}


Medical image segmentation \cite{wang2022medical} plays a crucial role in medical image analysis, which are usually trained in a fully-supervised manner. However, this kind of approach heavily suffers expensive annotation cost of providing pixel-level labels, impeding practical clinical application. In recent years, a wave of weakly-supervised segmentation models has emerged, which operate with different label levels, such as image-level\cite{hu2023conditional,wang2023boosting,xie2022weakly}, point-level\cite{gama2021learning,roth2021going}, scribble-level\cite{li2023scribblevc,wang2023s,wang2023weakly}, and bounding box-level\cite{groger2022boxshrink,wei2023weakpolyp,xu2021boxadapt} methods. By leveraging techniques such as reinforcement learning and active learning, they bridge the gap between pseudo-labels and ground truths, enabling pixel-level segmentation for medical images. Despite their innovative approaches, the current challenge lies in the fact that the results achieved by these methods still fall short of the performance exhibited by fully-supervised learning ways. Therefore, we aim to study lower-cost and higher-quality pseudo-labels for weakly-supervised medical image segmentation.

The Segment Anything Model (SAM)\cite{kirillov2023segment}, a general visual foundation segmentation model, has garnered widespread attention due to its remarkable segmentation and robust zero-shot generalization capabilities. Although SAM has been trained using large-scale data with pixel-level labels, its performance for medical image segmentation is unsatisfactory due to the lack of reliable clinical training data. Consequently, many researchers have fine-tuned SAM specifically in medical domains\cite{ma2024segment,ye2023sa,lei2023medlsam,deng2023sam}, including full fine-tuning and parameter-efficient tuning, achieving promising performance. Nonetheless, models based on SAM still require providing manual visual prompts (e.g. point and box prompts) for each image, increasing the difficulty and time required for expert physicians to make annotations. Therefore, we aim to explore a novel and automatic approach only using simple text cues to accomplish weakly-supervised medical image segmentation, through equipping SAM with a language-to-vision prompt converter. Moreover, to further conduct cross-modal fusion, we seek to better integrate text cues into the target visual segmentation model, ultimately enhancing the effect of language-driven segmentation performance.

In this paper, we propose a weakly-supervised medical image segmentation pipeline, \textbf{SimTxtSeg}. After the establishment of a domain-specific pre-training framework, the text prompt can easily be converted into a visual prompt and a pseudo-mask. Hence, only with a simple text cue, a target segmentation model can be trained in a weakly-supervised manner, eliminating the need of repeatedly providing pixel-level annotations. The most significant problem we investigated is how to effectively integrate information from simple text cues into the visual segmentation task model, such as transforming textual prompts into visual ones. Overall, we put forth SimTxtSeg, consisting of two key components: a Textual-to-Visual Cue Converter and a Text-Vision Hybrid Attention module.

\textbf{We highlight our contributions as follows:}
\textbf{1)} We propose to address weakly-supervised medical image segmentation using simple textual prompts, by extending the zero-shot generalization capability of SAM, thereby reducing the burden of pixel-level annotation on medical images.
\textbf{2)} The proposed SimTxtSeg includes a Textual-to-Visual Cue Converter(TVCC) and a Text-Vision Hybrid Attention(TVHA) module, promoting the integration of textual cues into the visual medical image segmentation task.
\textbf{3)} Through extensive comparison and ablation experiments, we validate the effectiveness of our approach, demonstrating state-of-the-art performance across multiple datasets, including scenarios such as intestinal polyp segmentation and MRI brain tumor segmentation.





\section{Proposed Method}
\begin{figure}[t]
\centering
\includegraphics[width=\textwidth]{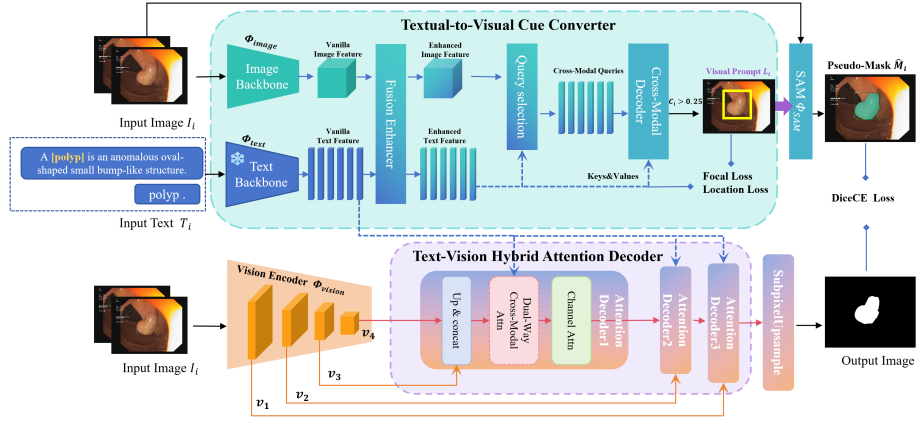}
\caption{\textbf{The framework of SimTxtSeg.} The textual-to-visual cue converter enables SAM to generate pseudo masks via text cues. Then, the weakly-supervised segmentation model is enhanced by text-vision hybrid attention.}
\label{fig:framework}
\end{figure}

\subsection{Problem Formulation}
Given a dataset with $N$ image-text pairs, $i.e.,D=\{(I_{1},T_{1}),..., (I_{N},T_{N})\}$, where $I_{i}\in \mathbb{R}^{H\times W\times 3}$ represents the $i$-th images, $H,W$ represent height,width, and $T_{i} \in \mathbb{R}^{L}$ is a brief description of the image, $L$ is the length of the sentence. First, we aim to train a textual-to-visual cue converter capable of directly localizing regions of interest in an image using simple descriptive text, then obtain the pseudo-masks $\hat{M} \in \mathbb{R}^{H\times W\times 1} $ by using SAM's zero-shot capability, thus eliminating the need for pixel-level annotations from doctors.
\begin{equation}
    B_{i},S_{i}=\Phi_{converter}(\Phi_{image}(I_{i}),\Phi_{text}(T_{i})),\label{e:convertor}
\end{equation}
\begin{equation}
    \hat{M}_{i}=\Phi_{SAM}(I_{i},B_{i}),\label{e:SAM}
\end{equation}
where $B_{i},S_{i}$ denote the bounding boxes and their confidence predicted by our textual-to-visual cue converter($\Phi_{converter}$), which has an image backbone ($\Phi_{image}$) and a text backbone ($\Phi_{text}$). $\Phi_{SAM}$ represents Segment Anything Model. Second, we propose a text-guided medical image segmentation model incorporated with a text-vision hybrid attention module in the decoder. To demonstrate the effectiveness of this weakly-supervised manner, we train it with image-text pairs $D_{train}$ and the pseudo-masks $\hat{M}_{train}$. The overall pipeline is shown in Fig. \ref{fig:framework}.


\label{sec:Problem Scenario}

\subsection{Textual-to-Visual Cue Converter}
Inspired by GroundingDINO\cite{liu2023grounding}, the construction of our Textual-to-Visual Cue Converter consists of image and text backbones for feature extraction, a feature enhancer for image and text feature fusion, a language-guided query selection module for query initialization, and a cross-modality decoder for box refinement. We utilize the mmdetection framework \cite{MMDetection} to fine-tune the Textual-to-Visual Cue Converter based on Swin-T \cite{liu2021swin}, employing a domain-specific medical dataset.  Our training data are transformed to the ODVG format for precise alignment of regions and phrases, $i.e.,D_{train}=\{(I_{1},T_{1},G_{1}),... ,(I_{N},T_{N},G_{N})\}$, where $G_{i}=\{\{Bbox_{1},Phrase_{1}\},..., \{Bbox_{N},Phrase_{N}\}\}$ contains the bounding boxes and their corresponding phrases, $N$ is the number of the lesions in the image. During training, we keep the weights of position embedding, backbone, and the language model (BERT-BASE) fixed, focusing solely on training the feature enhancer and cross-modality decoder. Consequently, it accurately generates precise annotation boxes based on textual cues.

Once the Textual-to-Visual Cue Converter is pre-trained, we can straightforwardly transfer the textual prompts into visual prompts for any new dataset within the same medical domain. Then we employ SAM as our pseudo-masks generator with the visual prompts $B$, configuring the confidence threshold at 0.25. In our study, both the vanilla SAM and SAM-med2d are experimented.
\label{sec:Pseudo-Masks Generation}

\subsection{Text-Guided Segmentation with Text-Vision Hybrid Attention}
\label{sec:txt-vision hybrid attention}
The objective of our work is to train a medical image segmentation model based on weakly-supervised text cues. Notably, these simple text cues serve a dual purpose: to generate pseudo-labels for supervision and to be directly integrated into the target segmentation model, effectively infusing intricate semantic details into the visual task model.

\textbf{Vision Encoder \& Text Encoder}: Given an image $I$, we choose ConvNext-Tiny\cite{Woo_2023_CVPR} as our vision encoder $\Phi_{vision}$: $v_{i}=\Phi_{vision}(I),$ 
where $i$ refers to the $i$-th layer in the backbone. We extract its first four layers of output for feature fusion, which are defined as $v_{1}\in \mathbb{R}^{H/4\times W/4\times C1}$, $v_{2}\in \mathbb{R}^{H/8\times W/8\times C2}$, $v_{3}\in \mathbb{R}^{H/16\times W/16\times C3}$, $v_{4}\in \mathbb{R}^{H/32\times W/32\times C4}$. Given a sentence $T$, we take BERT-BASE\cite{devlin2018bert} as our tokenizer and text backbone, and take its last embedding $t \in \mathbb{R}^{l\times C}$, where $l$ is the length of token, and $C$ refers to the feature dimension. $t=\Phi_{text}(tokenize(T)).$
    
\begin{figure}[t]
\centering
\includegraphics[width=\textwidth]{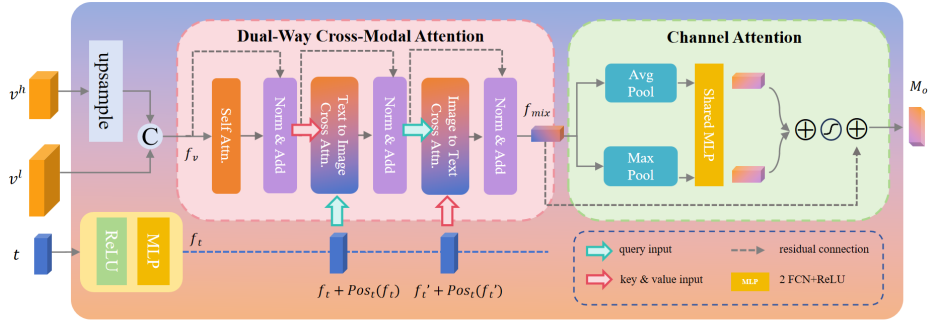}
\caption{Detailed structure of our text-vision hybrid attention decoder layer, containing the essential dual-way cross-modal attention and channel attention.}
\label{fig:attention}
\end{figure}

\textbf{Text-Vision Hybrid Attention Decoder}:
We employ three Text-Vision Hybrid Attention decoder layers, and a subpixel-upsample layer in our decoder. The details of Text-Vision Hybrid Attention decoder layer are illustrated in Fig. \ref{fig:attention}, which consists of a dual-way cross-modal attention and a channel attention. 

Let $v^{h}$ represents the high-level feature from the previous decoder layer ($v^{h}=v_4$ in the first decoder), and $v^l$ represents the low-level feature from the corresponding encoder layer, we upsample $v^h$ and concatenate it with $v^l$, then obtain the output $f_v$.

As a cue, the text embedding $t$ is aligned with visual feature dimensions through a projection layer, which is shown in the following equation: 
\begin{equation}
    f_t=\Phi_{projection}(t) = ReLU(\operatorname{MLP}(t)),
\end{equation}
where $\operatorname{MLP}$ contains a 1 $\times$ 1 convolution layer, a GELU activation function and a linear layer. $f_t \in \mathbb{R}^{L_j\times C_j}$ is the output text embedding, $L_j, C_j$ represent the length and channel number of the output token in the $j$-th decoder layer. Also, the image embedding's shape is projected into $HW \times C_j$, consistent with $f_t$.

For a more fine-grained integration of text and visual features, we propose \textbf{a dual-way cross-modal attention} in Fig. \ref{fig:attention}. Given $f_v,f_t$ representing aligned image and text embeddings, the dual-way cross-modal attention module performs three steps. 
First, we compute self-attention on the $f_{v}$, using image position embedding as the query and key, $f_{v}$ as value. Residual connection is employed to preserve the vision feature. The self attention is processed as: 
\begin{equation}
    f_v'=LayerNorm(MHSA(f_v))+f_v,
    \label{e:sa}
\end{equation}
where $MHSA(\cdot)$ refers to Multi-Head Self Attention.
Second, a text-to-vision attention is applied, which means cross-attention from text (text position embedding as query) to the image embbeding (image position as key, $f_{v}$ as value). 
Then, a norm-and-add layer is applied. The text-to-vision cross-attention process is shown in Eq. \ref{e:catv}: 
\begin{equation}
    f_t'=LayerNorm(MHCA_{tv}(f_t,f_v'))+f_t,
    \label{e:catv}
\end{equation}
where $MHCA_{tv}(\cdot)$ represents text-to-vision Multi-Head Cross-Attention.
Finally, we employ a vision-to-text attention, with image position embedding as query, text position as key, and $f_{t}$ as value, followed by the add-norm function, to get the fused feature $f_{mix}$:
\begin{equation}
    f_{mix}=LayerNorm(MHCA_{vt}(f_v',f_t'))+f_v',
\end{equation}
where $MHCA_{vt}(\cdot)$ represents vision-to-text Multi-Head Cross-Attention.


To further exploit the most useful feature channels, we introduce channel attention to automatically highlight the relevant feature channels while suppressing irrelevant channels. As is shown in Fig. \ref{fig:attention}, the mixed feature $f_{mix}$ undergoes global max pooling and global average pooling based on its width and height to fuse the spatial information across the entire feature map. The pooled features are individually processed through an MLP, which learns channel-specific weights and biases to enhance or suppress certain features. Then, the MLP outputs are element-wisely summed together and passed through a sigmoid activation, to obtain the decoder layer's output feature map $M_o$ as:
\begin{equation}
\label{e:ca}
\mathbf{M}_{\mathbf{o}} =Sigmoid(\operatorname{MLP}(\operatorname{AvgPool}(f_{mix}))+\operatorname{MLP}(\operatorname{MaxPool}(f_{mix}))) + f_{mix}.
\end{equation}

 
\section{Experiments and Results}

\subsection{Experiment Setup}


\noindent\textbf{Colonic Polyp Dataset:} We utilize the following datasets for colonic polyp segmentation:  CVC-ClinicDB\cite{bernal2015wm}, CVC-ColonDB\cite{tajbakhsh2015automated}, ETIS-LaribPolypDB\cite{silva2014toward}, \\Kvasir\cite{jha2020kvasir}, PolypGen\cite{ali2023multi}. In total, there are 3,784 images of colonic polyps, including both images appearing polyps and normal cases. We randomly split these datasets into training (3190 images), validation (299 images), and testing (295 images) set as the ratio of 8:1:1. The image size is reshaped to (384$\times$384).

\noindent\textbf{MRI Brain Tumor Dataset:} For brain tumor segmentation, we utilize LGG Segmentation Dataset\cite{buda2019association} from The Cancer Imaging Archive, which comprises 3,929 brain MRI images. Other settings remain consistent with the Polyp datasets.

\noindent\textbf{Text Cues:} We have designed two kind of text prompt granularities for each task: individual words and descriptive sentences. To avoid handcrafted prompting cost, we use GPT-4 to generate a concise sentence within 20 words. In the subsequent analysis, we will evaluate the effectiveness of these different granularities for SimTxtSeg.

\noindent\textbf{Evaluation Metrics:} We adopt mean Intersection over Union (mIoU) and mean Dice coefficient to evaluate the medical image segmentation performance.

\noindent\textbf{Implementation Details:}
To pre-train the Textual-to-Visual Cue Converter, we employed Adam optimizer with an initial learning rate of $2\times10^{-4}$, weight decay of $1\times10^{-4}$, a batch size of 8, and trained it for 100 epochs. As for the parameter scheduler, we adopted both LinearLR and MultiStepLR.
To train the text-guided segmentation model with Text-Vision Hybrid Attention, we freeze the text branch parameters and employ ConvNeXt as the vision backbone, with an input image size of 384. The learning rate adjustment strategy is ReduceLROnPlateau.
All the methods are implemented using PyTorch, accelerated by an NVIDIA 4090 Ti GPU. 

\subsection{Comparisons with the State-of-the-Art Methods}
\begin{figure}[t]
\centering
\includegraphics[width=\textwidth]{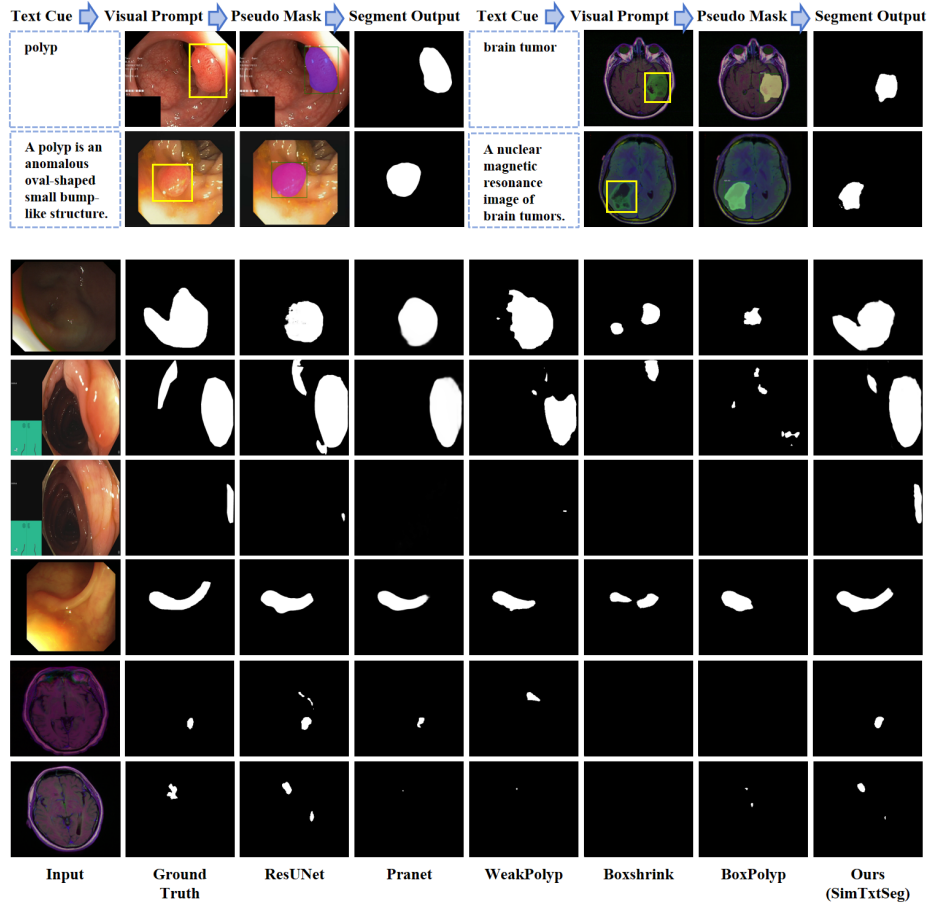}
\caption{Qualitative visualization on polyp and brain tumor segmentation.}
\label{fig:visualization}
\end{figure}

Comparison results against seven state-of-the-art methods are reported in Table~\ref{table:sota}.  These methods fall into two categories: three fully-supervised models (ResUNet\cite{diakogiannis2020resunet}, PraNet\cite{fan2020pranet}, and Ariadne’s Thread\cite{zhong2023ariadne}) and four weakly-supervised models with different label levels (WeakPolyp\cite{wei2023weakpolyp}, BoxPolyp\cite{wei2022boxpolyp}, Boxshrink\cite{groger2022boxshrink}, and S$^2$ME\cite{wang2023s}). We compared the segmentation performance of these SOTAs with our proposed pseudo-label generator(Pseudo-L:TVCC+SAM) and the final weakly-supervised model(SimTxtSeg-w-TVHA). It is observed that the generated pseudo-label quality is roughly on par with that of the SOTA fully-supervised models, with even a slight edge on the polyp dataset, and our final segmentation performance surpasses other SOTA weakly-supervised models. Among different kinds of weak supervision cue, our text-based cue is the weakest annotation without any spatial labeling and has the lowest cost compared to visual cues like boxes and scribbles which still cost a lot. On the polyp dataset, we achieve a +1.38\% improvement in mDice and a +3.36\% improvement in mIoU. Moreover, on the brain tumor dataset, our method achieves a +4.1\% improvement in mDice and a +3.94\% improvement in mIoU. Qualitative comparison of segmentation performance is visualized in Fig.~\ref{fig:visualization}.

\begin{table}[t]
\caption{Comparisons with state-of-the-arts on polyp and brain tumor segmentation, containing 3 fully-supervised and 4 weakly-supervised models. The results marked in gray means our generated pseudo-mask quality used for supervision.}
\resizebox{\textwidth}{!}
{
\begin{tabular}{ccc|cc|cc}
\hline
\multicolumn{3}{c|}{} & \multicolumn{2}{c|}{Polyp} & \multicolumn{2}{c}{Brain Tumor} \\
\multicolumn{3}{c|}{\multirow{-2}{*}{Method}} & mIoU(\%) & mDice(\%) & mIoU(\%) & mDice(\%) \\ \hline
\multicolumn{2}{c|}{} & ResUNet(2020) & 75.31 & 82.60 & 58.42 & 71.27 \\
\multicolumn{2}{c|}{} & PraNet(2020) & 81.32 & 87.30 & 74.14 & 82.49 \\
\multicolumn{2}{c|}{\multirow{-3}{*}{Fully-Supervised}} & Ariadne’s Thread(2023) & 80.65 & 87.14 & 71.55 & 81.20 \\ \hline
\multicolumn{1}{c|}{} & \multicolumn{1}{c|}{box} & WeakPolyp(2023) & 79.40 & 85.61 & 63.43 & 74.82 \\
\multicolumn{1}{c|}{} & \multicolumn{1}{c|}{scribble} & S2ME(2023) & 49.62 & 66.33 & 15.38 & 26.66 \\
\multicolumn{1}{c|}{} & \multicolumn{1}{c|}{ box+half anno. } & BoxPolyp(2022) & 79.11 & 86.86 & 67.40 & 77.64 \\
\multicolumn{1}{c|}{} & \multicolumn{1}{c|}{box} & boxshrink(2023) & 64.22 & 78.21 & 57.02 & 66.36 \\ \cline{2-7} 
\multicolumn{1}{c|}{} & \multicolumn{1}{c|}{-} &\cellcolor{mygray} Pseudo-L:TVCC+SAM & \cellcolor{mygray}81.06 & \cellcolor{mygray}87.46 & \cellcolor{mygray}72.38 & \cellcolor{mygray}81.69 \\
\multicolumn{1}{c|}{} & \multicolumn{1}{c|}{-} & SimTxtSeg-w/o-TVHA & 74.92 & 83.15 & 66.57 & 77.86 \\
\multicolumn{1}{c|}{} & \multicolumn{1}{c|}{-} & SimTxtSeg-w/o-CMA & 80.83 & 87.22 & 71.16 & 81.57 \\
\multicolumn{1}{c|}{} & \multicolumn{1}{c|}{text} & SimTxtSeg-w/o-CA & 80.64 & 86.87 & 70.42 & 80.97 \\
\multicolumn{1}{c|}{\multirow{-7}{*}{\makecell[c]{  Weakly-\\ Supervised  }}} & 
\multicolumn{1}{c|}{text} & SimTxtSeg-w-TVHA & \textbf{82.47} & \textbf{88.24} & \textbf{71.34} & \textbf{81.74} \\ \hline
\end{tabular}
}
\label{table:sota}
\end{table}

\begin{table}[t]
\caption{Eval. of prompt types and SAM variants for pseudo-mask generation.}
    \centering
    \scalebox{0.94}{
    \begin{subtable}[t]{0.495\linewidth}
        \begin{tabular}{c|cccc}
\hline
    \multirow{2}{*}{\makecell[c]{ Prompt\\ Type }}& \multicolumn{2}{c}{Polyp } & \multicolumn{2}{c}{ Brain Tumor} \\

 &  mIoU & mDice  &  mIoU & mDice \\ \hline
\cite{liu2023grounding}-w-class & 22.15 & 29.29 & 10.89 & 13.02 \\
Class name & 80.84 & 87.30 & 68.00 & 78.15 \\
Sentence & \textbf{81.06} & \textbf{87.46} & \textbf{68.30} & \textbf{78.36} \\
 \hline
\end{tabular}
    \end{subtable}
    \begin{subtable}[t]{0.495\linewidth}
       \begin{tabular}{c|cccc}
\hline
\multirow{2}{*}{ SAM Variant } & \multicolumn{2}{c}{Polyp } & \multicolumn{2}{c}{ Brain Tumor} \\

 &  mIoU & mDice  &  mIoU & mDice \\ \hline
SAM-base & 76.87 & 84.32  & \textbf{72.38} & \textbf{81.69} \\
SAM-huge & \textbf{81.06} & \textbf{87.46} & 68.30 & 78.36 \\
SAM-med2d-base  & 70.62 & 79.07 & 67.20 & 77.34 \\
 \hline
\end{tabular}
    \end{subtable}
    }
    \label{tab:array}
\end{table}

\subsection{Ablation Study}
\noindent\textbf{Impact of prompt types.} We evaluated class name and sentence as text cues during training textual-to-visual cue converter and compared their effectiveness for pseudo-mask generation on SAM-huge. Also, we tested the performance by the original GroundingDINO\cite{liu2023grounding} with class name prompt. As Table~\ref{tab:array} shows, training the textual-to-visual cue converter with sentences (e.g. A polyp is an anomalous oval-shaped small bump-like structure.) tends to yield slightly better results than training it with class names (e.g. polyp), since the converter generates better box cues, achieving 80.1\% mAP for polyp and 74.8\% mAP for brain tumor. The original GroundingDINO fails to generate useful pseudo masks. 

\noindent\textbf{Impact of SAM variants.} We compared three pre-trained SAM models: SAM-huge, SAM-base, and SAM-Med2d-base\cite{ye2023sa} for pseudo-label generation, which differ in model parameters and pretraining dataset. As seen in Table~\ref{tab:array}, the SAM-huge performs better for polyp images while SAM-base yields superior results for brain tumor dataset. Due to extensive pretraining of SAM-Med2d-base specifically with CT and MRI data, it exhibits significant bias when applied to polyp data, resulting in poor generalizability compared to the general SAM.

\noindent\textbf{Impact of our TVHA.} SimTxtSeg-w/o-TVHA denotes the model without Text-Vision Hybrid Attention, using UNet decoder instead. SimTxtSeg-w/o-CMA denotes the model without Dual-Way Cross-Modal Attention, SimTxtSeg-w/o-CA denotes the model without Channel Attention. From Table~\ref{table:sota}, it is observed that, after incorporating the TVHA, our model’s performance has significantly improved. Specifically, on the polyp dataset, we observe a +5.09\% increase in mDice and a +7.55\% increase in mIoU. The contribution of both modules to model performance improvement is roughly equal, but using them together achieves the best results. Also, we surprisingly find that the results by SimTxtSeg even surpass the pseudo masks by TVCC+SAM used for weakly supervision.



\section{Conclusion}

This paper proposes an effective SimTxtSeg for weakly-supervised medical image segmentation via inputting simple text cues, which contains a textual-to-visual cue converter and a text-vision hybrid attention mechanism. Extensive experiments are conducted to prove that, using simple text cues, our approach achieves state-of-the-art performance with minimal supervision. In the future, we will extend our method to more medical image analysis areas and fuse the TVCC and SAM into an end-to-end fashion for improvement.



\begin{credits}
\subsubsection{\ackname} This work was partially supported by the National Natural Science Foundation of China (Grants No 62106043, 62172228), and the Natural Science Foundation of Jiangsu Province (Grants No BK20210225).

\subsubsection{\discintname}
The authors have no competing interests to declare that are
relevant to the content of this article.
\end{credits}
%
%
%
\bibliography{main.bbl}
\bibliographystyle{splncs04}
%



\end{document}